\documentclass[11pt]{amsart}
\usepackage{graphicx}
\usepackage{amsfonts, amsmath, amsfonts, amssymb}
\newtheorem{thm}{Theorem}[section]
\newtheorem{cor}[thm]{Corollary}
\newtheorem{lem}[thm]{Lemma}
\newtheorem{prop}[thm]{Proposition}
\theoremstyle{definition}
\newtheorem{defn}[thm]{Definition}
\newtheorem{exmp}[thm]{Example}
\theoremstyle{remark}
\newtheorem{rem}[thm]{Remark}
\numberwithin{equation}{section} \theoremstyle{quest}
\newtheorem{quest}[]{Question}
\numberwithin{equation}{section} \theoremstyle{prob}

\numberwithin{equation}{section} \theoremstyle{answer}

\numberwithin{equation}{section}

\begin{document}

\title[]{Isometric Embeddings in Imaging and 
Vision: Facts and Fiction}%
\author{Emil Saucan}%
\address{Department of Mathematics, Technion, Haifa, Israel}%
\email{semil@tx.technion.ac.il}%

\thanks{Research supported by the Israel Science Foundation Grant 666/06 and by European Research Council under the European Community's Seventh Framework Programme
(FP7/2007-2013) / ERC grant agreement n${\rm ^o}$ [203134].}%
\keywords{Nash's Theorem, $PL$-isometric embedding, Burago-Zalgaller construction, quasiconformal mapping, maximal dilatation}%

\date{\today}

\begin{abstract}
We explore the practicability of Nash's Embedding Theorem in vision and imaging sciences. In particular, we investigate the relevance of a result of Burago and Zalgaller regarding the existence of isometric embeddings of polyhedral surfaces in $\mathbb{R}^3$ and we show that their proof does not  extended directly to higher dimensions.
\end{abstract}


\maketitle

\section{Introduction}

Recently, the attention of the vision community was drawn towards
the important problem of isometric embeddings of manifolds in
$\mathbb{R}^n$ (\cite{kms}), in particular to the classical case of
surfaces and their embedding in $\mathbb{R}^3$ or $\mathbb{R}^5$
(\cite{bbk}). This approach was further extended by the employment
of metric techniques that achieved fame through the celebrated work
of Michael Gromov \cite{gr2} -- see, \cite{bbk}, \cite{bbk1},
amongst others. We do not consider here the critique of the metric
method in general, since that would be, indeed, against our belief that it represents a basic, essential tool in imaging (see, for reference, \cite{sa0}), but concentrate rather on the first problem: that of the
isometric embedding.

Let us begin by noting that, certainly the departure from the entrenched
view of surfaces as given only locally by parameterizations (usually
via by spline functions) has merit, as well as the realization that
surfaces, that is 2-dimensional manifolds need often 
to be embedded, in a
metrically controlled manner, in some $\mathbb{R}^N$, the value of
$N$ being, preferably, equal to 3. However, there are some deep and
disturbing problems stemming from this approach, that is based upon
the celebrated Nash Embedding Theorem.

We begin with the least of these problems: the feasibility of
finding an isometric embedding of a given, smooth orientable surface (or, more generally, an orientable manifold) in some
$\mathbb{R}^N$, for $N$ large enough. The root of the difficulty in
writing and implementing an algorithm based upon Nash's Theorem
resides in the fact that this theorem\footnote{together with other
two recently famous methods: the circle packing (\cite{hbssrsr},
\cite{hs}) and the Riemann Mapping Theorem (see, e.g. \cite{bea}).}
is a obtained via a fixed point method. The
impediment 
here resides not only in the fact that such a method
requires, at least theoretically, an infinite number of
steps\footnote{ and that it is far less algorithmic in nature than
the Picard Fixed Point Theorem of Differential Equations
fame}, but rather in the disturbing fact that the manifolds
in the approximating sequence are not usually
submanifolds of the same $\mathbb{R}^N$, where $N$ represents the dimension
of the Euclidean space
in which the target (approximated)
manifold, is (ideally) to be embedded -- see also Remark 1.1 below.\footnote{A very similar problem arises also
in the computational Riemann Mapping Theorem.} 
This is not just a theoretical, quasi-philosophical quandary, but rather a serious
impediment: indeed, since computers, by their very nature, can
perform only a finite number of approximations, the computed
manifold will be embedded in a dimension different from the one
stipulated (and often depicted -- see also Remark \ref{rem:mistake} below).

Even if we can, somehow, surmount this difficulty, one still is
faced with the dire specter of dimensionality. To make this
assertion clearer (and more concrete) let us recall a few facts
regarding 
Nash's Theorem:

The celebrated Nash embedding theorem assures the existence
of an isometric embedding of any $\mathcal{C}^k, (3 \leq k \leq \infty)$ orientable\footnote{In the following, all manifolds are supposed to be orientable, except if otherwise specifically stated.}
manifold of
dimension $n$ into some $\mathbb{R}^N$, for some $N$ sufficiently large. In contrast
with the topological case, were one has the classical Whitney's Theorem (see Appendix 1),
the embedding dimension ensured by the Nash theorem is prohibitively
large, 
 giving $N= 
 \frac{n(3n + 11)}{2}$. 
(Note, however, that in Nash's original work, the embedding dimension for noncompact manifolds was $N = \frac{3}{2}(n^2 + 5n)(n+1)$.)

Even with the further dimension reductions of Gromov (\cite{gr1},
\cite{gr}) and G\"{u}nter (\cite{gu}, \cite{gu1}), the assured embedding dimension, namely $(n^2 +
10n +3)$, for $n \geq 3$, and $(n + 2)(n +3)$, if $n\geq 4$ and
($\max{\{\frac{n(n + 3)}{2} + 5,\frac{n(n +5)}{2}\}}$, respectively, numerics are not very 
promising: the embedding dimension $N$ of a surface (i.e.
2-dimensional manifold) provided 
 by the original Nash Theorem is 17, and by Gromov's  and  G\"{u}nter's improvements, $10$. However,
a special method developed by Gromov (\cite{gr}, p. 298) decreases the
embedding dimension for compact surfaces to $5$, while for compact 3-manifolds the lowest guaranteed embedding dimension is $N = 13$ -- see \cite{gr}, p. 305. (Moreover, there exists a local isometric embedding of a given $M^2$ in any 5-dimensional manifold $N^5$ -- see \cite{gr}.)

Strikingly, there is nothing really known about the case $k = 2$
(its omission in the discussion above being no mistake).\footnote{it is not even known whether $\mathcal{C}^2$ manifolds admit $\mathcal{C}^2$ isometric immersions in $\mathbb{R}^3$}
Since curvature (of differentiable surfaces, at least) is essentially a $\mathcal{C}^2$ notion\footnote{Indeed, almost all the corpus of classical differential geometry of surfaces may be developed
assuming only this degree of smoothness.} one would count on a general Nash embedding theorem for this case, that would allow a straightforward application in imaging.

When one further relaxes 
the smoothness condition, the embedding dimension
decreases dramatically: any $\mathcal{C}^1$ orientable 2-manifold is
isometrically embeddable in $\mathbb{R}^3$ (see \cite{na1}, \cite{ku}).\footnote{The $PL$ version of this result (and its extension to higher dimensions) represents the subject of following 3 sections.}
Indeed, since the proper (i.e. differentiable)
notion of curvature makes sense only for manifolds of class $\geq
2$, the additional 
dimensions that are required to deal with 
curvature,\footnote{using, for instance, the Gauss Equation -- see, e.g. \cite{a}, \cite{D'A}}
are not necessary in the $\mathcal{C}^1$
case. The role of the lack of differentiability is further
emphasized by the fact that by Nash-Kuiper Theorem, even the {\em flat}
torus is $\mathcal{C}^1$ isometrically embeddable in $\mathbb{R}^3$,
in contrast with Tompkins' Theorem \cite{to} (see also \cite{spi5},
pp. 196-197) that asserts that compact flat $\mathcal{C}^2$
manifolds are not even isometrically immersable into
$\mathbb{R}^{2n-1}$. Not only this, but also the following result (due to Kuiper \cite{ku}) holds: The unit sphere $\mathbb{S}^n \subset \mathbb{R}^{n+1}$ admits an isometric $\mathcal{C}^1$ immersion in $\mathbb{R}^{n+1}$, for any $n \geq 1$. (This being in sharp contrast with the fact that any such $\mathcal{C}^2$ immersion is congruent to the unit sphere, for $n \geq 2$.)

Moreover, any $\mathcal{C}^0$ embedding is smoothable to $\mathcal{C}^1$, therefore Whitney's Embedding Theorem (see Appendix 1) implies Nash's Embedding Theorem (for $n \geq 2$).

We should note that the case of analytic manifolds is, again, different: In \cite{na3} Nash proved that any compact (real) analytic n-dimensional manifolds has an isometric embedding in $\frac{3}{2}(n^2+5n)$. Gromov \cite{gr1} extended Nash's result to include noncompact manifolds and also reduced the embedding dimension (for both cases) to $\frac{1}{2}(n^2 + 7n +10)$.

\begin{rem}
The striking difference between the results for the various degrees of smoothness emphasize once more the delicate manner in which one should approach the embedding problem, and even more so its practical applications. Moreover, even small variations of the metric, especially those producing change in the sign of (Gaussian) curvature (see also Remark \ref{rem:Ck-loc}), can abruptly change\footnote{and, unfortunately, usually increase} the embedding dimension (see \cite{HH} for a plethora 
of results in this direction).

It follows that, when using approximating sequences of $PL$ manifolds and the isometric embedding technique, one can not ascertain with any degree of certainty that the sequence of embeddings remains in the same (minimal) dimension. Indeed, those familiar with polygonal meshes (e.g. people working in the field of graphics) know that -- quite counterintuitively -- even polygonal approximations of spheres have vertices of (concentrated) negative curvature (that is saddle points).


Therefore, the only general assured embedding dimension is that guaranteed by the Nash-Gromov-G\"{u}nther Theorem.
\end{rem}

\begin{rem}
Since we perceive shape, rather than distance, and since, by the previous remark (and the discussion preceding it, as well as by Remark 3.5 below), curvature, hence shape, is lost, it follows that $\mathcal{C}^1$ isometric embeddings and their $PL$ counterparts are far less useful in imaging, recognition and matching purposes than hoped for.
\end{rem}

Before proceeding any further, we have to add that we are aware that some readers are less familiar with some of the necessary background in differential topology and that, in any case, it would be best to refresh the basic 
necessary notions regarding immersions and embeddings. Therefore, we have included a glossary of relevant notions as Appendix 1. Given its goal, the pace is slow and the  tone is rather didactical, therefore many a reader may want to omit it. However, since we also discuss therein one of the common misunderstandings regarding isometric embeddings, we believe it may be useful to all.


\section{A suggested solution: PL isometric embeddings}

It has been suggested to us \cite{po} that the disquieting facts
regarding the smooth embeddings considered in Nash's Theorem need
not disturb us to much, for, in the imaging and graphics practice, one is always faced with
$PL$-flat surfaces (``triangular meshes'') and for these 
a highly surprising and widely unexpected
result exists\footnote{For some more recent, seemingly paradoxical, related results, see e.g. \cite{BPS}, \cite{Pak1}, \cite{pak2}.}, namely the following theorem due to Burago and Zalgaller:

\begin{thm}
Any compact orientable $PL$ $2$-manifold admits an isometric embedding in
$\mathbb{R}^3$.
\end{thm}

The common wisdom regarding the statement above is, of course, that in imaging and graphics such surfaces represent the geometric object under investigation, or at least a ``decent'' approximation of it.

Unfortunately, this does not represent the solution of the problem
in question. Indeed, a number of problems arise as soon as one examines this theorem a bit closer.

To begin with, the formulation above, while convenient and easy to recall, is not the correct one. The correct one can be found in \cite{gr}, p. 213 (but recall also the title of the paper  \cite{bz} of Burago and Zalgaller):

\begin{thm}[Burago-Zalgaller \cite{gr}]
Every compact oriented surfaces with a piecewise linear metric can be piecewise linearly isometrically embedded in $\mathbb{R}^3$.
\end{thm}

First thing that strikes us is the apparently cumbersome and futile 
new terminology. However - as usually 
is the case in Mathematics, these apparent pointless minutiae and stresses are essential, and not due to just a whim of the mathematician. To comprehend this better in the case at hand, we should first understand  
the difference between the two (apparently identical) notions:\footnote{Given the space limitations and the desire for cohesiveness, we must assume the reader is familiar with the very basic notions of $PL$ topology.
(For a deep, yet enjoyable and not overly technical source on these notions, see \cite{th}. See also Appendix 1 for additional material on embeddings.)}

{\it $PL$ isometric embedding}
If $P$ is a simplicial polyhedron of dimension $n$, any simplicial map $f:P \rightarrow \mathbb{R}^m$ (which is, by definition, linear on any simplex of $P$) induces a flat metric on each such simplex and, in consequence, a (singular) Riemannian metric $g$ on $P$. More precisely, if $P$ has $k$ edges, then $g$ is uniquely determined by the the vector $(g_1,\ldots,g_k)$, where $g_i = \big({\rm length}(e_i)\big)^2$.
Saying that $f$ is a $PL$ isometric embedding means that $f$ is as above and, in addition, it is also an embedding.

{\it $PL$ isometric embedding of subdivided polyhedra}
In this case, the mapping (embedding) $f:P \rightarrow \mathbb{R}^m$ is required to be linearly isometric on the simplices of a simplicial subdivision of $P$. Evidently, by a sufficient number (albeit practically infinite) number of subdivisions, one can approximate the Riemannian case using
$PL$ metrics, i.e. such that each simplex of $P$ is isometric to a Euclidean simplex (in $\mathbb{R}^n$).

So, why is the first definition not 
adequate? 
The main problem is its rigidity: Informally put, one ``has to work with what he's got''. That is, further subdivisions (hence approximations) are not allowed. Therefore, this approach rapidly reduces to a largely combinatorial problem, at least in many of its aspects (see, e.g. \cite{Ka}).

Moreover (and more important) this rigidity is not just of convenience (so to say), quite the contrary -- it is essential. Indeed, most\footnote{Here ``most'' has a precise mathematical meaning: More precisely, {\it generic} simplicial mappings are {\it rigid}, 
where ``generic'' is a rather technical term (see \cite{gr}). }
$PL$ isometric maps are rigid, in the (geo-)metric sense:

\begin{thm} [\cite{gr}]
Every small deformation of a $n$-dimensional polyhedron embedded (immersed) in $\mathbb{R}^{n+1}$ is an isometry.
\end{thm}

In fact, the result just mentioned is more general, 
but to avoid a further detour, we refer to \cite{gr}. pp. 210-211.)



Obviously, the second notion is far more attractive, both for the geometer/analyst as well as for Computer Graphics and related fields. However, caution should be taken, since this is still a very flexible notion, since just the metric is to be preserved. For instance, one has the following result of Zalgaller (\cite{Za}):

\begin{thm}[Zalgaller, 1958]
Let $P$ be a simplicial polyhedron of dimension $n, n \leq 4$, endowed with a $PL$ metric. Then $P$ admits an {\it equidimensional} $PL$ map into $\mathbb{R}^n$.
\end{thm}

This is a very surprising and counterintuitive result.\footnote{However, it represents the $PL$ version of the $\mathcal{C}^1$ version (that holds for any $n$) -- see, e.g. \cite{gr}.} So, even though the embedding condition is omitted, it prepares us to understand somewhat better the problematic nature of the definition and of the Burago-Zalgaller theorem.

To further elucidate these notions 
we take below a closer look 
 at Burago and Zalgaller's proof.


\section{The Burago-Zalgaller construction}

Main idea of the proof is -- not very surprisingly -- to adapt the proof of the $\mathcal{C}^1$ Nash-
Kuiper Embedding Theorem.\footnote{We preserve in the following overview of the proof 
the notation of \cite{bz}. For copyright reasons we do not reproduce, however, the figures included therein, but rather refer the reader to the original source.}
For this, one starts with a (smooth) embedding of the given polyhedron
in $\mathbb{R}^3$, composed with a contracting homotety. Then:

\begin{itemize}

\item Carry out a sequence of stages, divided in turn into a large
number of steps, each of which improves the approximation to isometry and such that the function obtained at each stage is 
{\it short}:

\begin{defn}({\it Short mappings})
Let $(X,d)$ and $(Y,\rho)$ be metric spaces. A map $f:X \rightarrow Y$ is called $C$-{\it short} iff

\[\rho(f(x),f(y)) \leq Cd(x,y)\,, {\rm for\; all\;} x,y \in X\,.\]

$f$ is called {\it short} (or {\it contracting}) iff it is $C$-short, for some $C < 1$.
\end{defn}

\item Add ``ripples'', producing thus a $PL$ version of Kuiper's adaptation \cite{ku} of {\it Nash's
 twist} \cite{na1}) such that one will have ``enough space'' to isometrically embedded the surface in $\mathbb{R}^3$.

\end{itemize}

We won't dwell too much in the details of the proof, just mention some of the principal ``geometric'' stages:

\begin{enumerate}
\item {\em Basic Construction Element}
      \begin{enumerate}
      \item Let $T = \triangle(A_1,A_2,A_3)$ and $t =
 \triangle(a_1,a_2,a_3)$
be acute triangles;
      \item let $B,b$ and $R,r$ the centers and radii of their
respective circumscribed circles;
      \item let $E_p = \frac{1}{2}A_kA_l, e_p = \frac{1}{2}a_ka_l;\, p,k,l \in
\{1,2,3\}$;
      \item and let $H_p = BE_p, h_p = be_p$.

Moreover, let $T \simeq t, A_kA_l > a_ka_l, k,l \in \{1,2,3\}$.

Then $T$ can be isometrically $PL$ embedded in
$\mathbb{R}^{3}$, as the pleated surface included in the right prism with base
$t$, such that $A_kA_lA_p$ fits $a_kE'_pa_lE'_ka_pE'_l$, where: $B'b  \bot t, B'a_p = R$ and $E'_p, E'_k, E'_l$ on the faces of the
prism, such that $a_kE'_p = Ei_pa_l = \frac{1}{2}A_kA_l$.
%

The following variations of the basic construction above are also considered:
               \begin{enumerate}
               \item Each angle $\varphi$ of $T$ satisfies the condition
$0 < \alpha < \varphi$ and $C\cdot A_kA_l > a_ka_l\,, C < 1$. Moreover,
$A_kA_l/a_ka_l \approx 1$.
               \item Each of the lateral faces of the prism -- including the broken lines
$a_kE'_pa_l$ -- can be (independently) slightly rotated around the lines
$a_ka_l$ such that the construction still can be performed. (The rotation angle depends upon the
constants $\alpha$ and $C$ above.)
               \end{enumerate}

               (In general, one has to simultaneously construct a large number of the units above.)
       \end{enumerate}

\item {\it Standard embedding near vertices}

Use the {\it standard conformal map} (or {\it folding}) from $K(\theta,\rho) =
\{0 \leq \varphi \leq \theta, \rho > 0\}$ to $K(\lambda,r) = \{0 \leq \psi \leq \lambda, r >
0\}$ given by:

\[\psi = \frac{\lambda}{\theta}\varphi,\; r = a\rho^{\lambda/\theta}.\]

(The most important case for our purposes being: $\lambda = 2\pi$.)

\item{\it The Triangulation and its refinement}

Let $A$ be a vertex of total angle $\theta$.

    \begin{enumerate}
    \item If $\theta < 2\pi$, then encircle $A$ by a small ``regular'' hexagon composed of $6$ triangles of apex angle $\theta/6$.

          Some small enough neighbourhood of $A$ the will be mapped by the standard conformal mapping onto a planar disk.

          Over each triangle included in such a neighbourhood, one can perform the basic construction, obtaining a $PL$ isometric embedding of this neighbourhood.

     \item If $\theta > 2\pi$, proceed analogously to the previous case but

           \begin{enumerate}
           \item In a small circular neighbourhood of radius $r_1$ map (a) isometrically on radial
segments and (b) using a $\theta/2pi$ contraction on circles centered at $A$;
           \item In a annular neighbourhood $\{r_1 < r < r_2\}$ use the standard conformal mapping with
the same contraction factor $\theta/2pi$.
           \end{enumerate}
     Replace the neigbourhood above with a ``cogwheel'' (i.e. a circle surrounded by isosceles ``triangles'' of sides, e.g. $2\delta$, and having as bases arcs of the same length).
     %
     %
     The interior of each ``cogwheel'' is $PL$ isometric embedded using ``ripples''. (The basic element of each such ``ripple'' is a pair of congruent triangles, having a common vertex in the center of the ``cogwheel'', one side (of each) being a radius, and a second common vertex built over the midpoint of an arc used in the construction of the ``cogwheel'' -- see Figure 4 of \cite{bz}).
     %
     %
     Away from neighbourhoods of vertices, refine the triangulation using only acute triangles.
     In particular, at convex vertices subdivide each triangle into $n^2$ similar triangles, for
some large enough $n$; while at non-convex vertices into almost regular triangles.
     %
     %

     \end{enumerate}
\end{enumerate}


\begin{rem}
Adaptations \cite{bz} of the main technique exposed above
ensure the existence of $PL$ isometric embeddings of (orientable) $PL$ manifolds with boundary and of $PL$ immersions of nonorientable $PL$ manifolds.
\end{rem}

\begin{rem}
A close examination of the arguments of the
proof shows that Theorem 2.2 can be extended to include (orientable)
non-compact manifolds with bounded (generalized) principal curvatures \cite{s}.
\end{rem}

Yet one naturally has to ask himself the following 

\begin{quest}
Is the Burago-Zalgaller Theorem applicable for Image Processing/Computer Graphics?
\end{quest}

Unfortunately, the answer is negative, for the following reasons:

\begin{enumerate}
\item The construction yields an (infinite) approximation process, akin to the original
Nash-Kuiper method, hence numerical errors have to contended with and taken into account.

\item The geometry\footnote{all important in any practical implementation} of the
limiting object is very far from the one of the ``target surface'': Not only is the resulting $PL$ surface strongly ``corrugated'' (as evident from the construction), it my also contain ``superfluous'' vertices, i.e. where the curvature (of the metric) is zero. Moreover, for surfaces of positive {\it extrinsic curvature}, it is quite possible that the surface admits not even an isometric immersion in $\mathbb{R}^3$ such that the extrinsic curvature equals the {\it intrinsic} one. (For the technical definitions and a simple example of a $PL$ $2$-sphere exhibiting this behavior, see \cite{BS}, p. 76.

Also, note that ``accidents''
in the original embedding can produce widely diverging subdivision schemes -- see also the comment on page 2 above, as well as Remark 1.1. (To grasp this widely divergent behavior, one should consider, for instance, the examples quoted in footnote 4, on page 2.)

\begin{exmp}({\it Burago-Zalgaller, Example 1.5})
For any $\varepsilon > 0$, the flat torus $\mathbb{T}^2$ (i.e. the ``topologist's torus'', obtained by ``gluing'' the opposites sides of a plane square via 
Euclidean translations) admits a
$PL$ isometrical embedding $\varepsilon$-close to the rotation (``round'')
torus (see, e.g. \cite{doC}, pp. 434-435).\footnote{Again, this result represents the $PL$ equivalent of its $\mathcal{C}^1$ counterpart -- see page 3 above.}
\end{exmp}

\begin{rem} \label{rem:curvature}
It is contended in \cite{BH}, p. 618, that the the Burago-Zalgaller embedding method 
preserves curvature. However, this assertion is not made in \cite{bz}.\footnote{In fact, the authors of \cite{bz} explicitly state (see \cite{bz}, p. 370) that they ``... prove the ... discrete analog of the well-known result of J. Nash and N. Kuiper on {\em $\mathcal{C}^1$-smooth isometric immersions}'' (our emphasis).}
 Indeed, this is not possible, as the example above clearly hints and as we shall explain in some detail below.

First, we should understand what type of curvature is preserved. Evidently, not the canonical (``smooth'') one of classical differential geometry, since the considered surfaces are not even $\mathcal{C}^1$. It may be that the authors of \cite{BH} refer to the fact that a piecewise flat surface (or polyhedron) remains piecewise flat during the embedding process and, as such, its curvature is identically zero at all the points that do not belong to the vertices and edges of the triangulation. However, it is explicitly emphasized in the very introduction of \cite{bz} that ``the metric of a polyhedron is locally flat {\em except}\footnote{our emphasis} at a finite collection of points; these points are the ``true'' vertices.''\footnote{\cite{bz}, p. 369}

However, it is precisely at these points that Gaussian curvature is concentrated (being the defect of the planar angles (of the faces) incident at any such vertex). This is a known, in fact, since Descartes, but it was introduced in modern Mathematics by Hilbert and Cohn-Vossen \cite{HC-V}, and developed first by Polya \cite{Pol} and then by Banchoff \cite{ban1}, \cite{ban2} and, more recently by Stone \cite{St} and Fu \cite{Fu}.%
\footnote{This definition of Gaussian curvature for polyhedral surfaces also facilitates an easy proof of the fact that no $PL$ isometric embedding of the flat torus, preserving discrete curvature, is possible in $\mathbb{R}^3$. This proof follows closely the one for the smooth, classical case (see e.g. \cite{doC}): Since the given polyhedral embedding $\mathcal{P}$ is compact, there exists a number $R > 0$ such that $\mathcal{P}$ is included in the interior of $\mathbb{S}^2_R(0)$ -- the sphere centered at the origin and of radius $R$. Let $R$ decrease continuously and let $R_0$ be the radius for which the sphere and the polyhedron have the first non-void intersection. Moreover, by elementary arguments, any such intersection point must be a vertex of the polyhedron. Since $\mathcal{P}$ is contained in the closure of $\mathbb{S}^2_{R_0}(0)$, that is on one side of the surface of the sphere, it follows that the sum of the angles at an intersection vertex $v_0$ must be smaller than $2\pi$, that is the discrete Gaussian curvature $K_{v_0} > 0$, in contradiction to the supposition that $\mathcal{P}$ is a realization of the flat torus, i.e. having discrete Gaussian curvature equal to zero at all its vertices.}
 (Mean curvature is, by contrast, concentrated along the edges of a polyhedral mesh, as the {\it dihedral angle} of the two faces who's intersection is any specific edge -- see \cite{LSE} for a succinct presentation and for the bibliography within.)

\end{rem}

\end{enumerate}

\section{A shattered hope}

Having seen that the Burago-Zalgaller construction is not applicable as such, one at least hopes for a positive answer to the following natural

\begin{quest}
Does Burago-Zallgaler's Theorem hold in dimension $n \geq  3$?
\end{quest}

Perhaps unexpectedly, and contrary to the unsubstantiated 
statement of \cite{BH},\footnote{In fact, the abstract \cite{KBP} does state the result but does not sustain 
it was actually obtained.}
the answer to this question is not
known! However, there are indications that the answer is negative.
These indications emerge from the proof in dimension $2$:

\begin{enumerate}

\item The proof is based on the previous result of Burago and
Zallgaler on the existence of acute triangulations.

Strangely enough, next to nothing is known about the existence of
such triangulations in dimension $n \geq 3$.\footnote{The little existing information is summarized by 
Zamfirescu \cite{Za}. (Some additional hope stems from a different
method developed recently by Tasmuratov \cite{Ta}.)}

\item The proof heavily relies on the use of the use of the standard conformal map to produce a mapping that (around
the vertices) is arbitrarily close to conformality (and, in the
end) to isometry.

However, in dimension $n \geq 3$ this is not possible: the analogue of the standard conformal map has dilatation bounded
away from $1$! Indeed, we can be more specific. But first, a
few technical\footnote{But hopefully not too technical -- we shall restrict ourselves to the simplest case.} details:

\begin{defn}({\it Wedges})
Let $x \in \mathbb{R}^n$ be a point with cylindrical coordinates $x =
(r\cos\varphi,r\sin\varphi,z_1,\ldots , z_{n-2})$. The set $D_\alpha = \{0 < \varphi < \alpha\}$,
$(0 < \alpha \leq 2\pi)$ is called a {\it wedge} of
angle $\alpha$.
\end{defn}

\begin{defn}({\it Foldings})
The homeomorphism. $f:D_\alpha \rightarrow D_\beta$, $f(r,\varphi,z) = (r,\frac{\alpha}{\beta}\varphi,z)$, $z = (z_1,\ldots , z_{n-2})$ is called
a {\em folding}.
\end{defn}

Before proceeding further, the reader should familiarize herself/himself with the technical notions regarding quasiconformal mappings. Not wishing to interrupt the flow of geometric arguments, we have concentrated these in a short appendix (Appendix 2).

\begin{prop}[Gehring-V\"{a}is\"{a}l\"{a} \cite{gv},\cite{v}] \label{prop:dist-coef1}
Let $D_\alpha,D_\beta$ be wedges, and let $f:D_\alpha \rightarrow D_\beta$ be the respective folding.
If $\alpha \leq \beta$, then $f$ is quasiconformal, with dilatations $K_I(f) = \frac{\alpha}{\beta}, K_O(f) \geq
(\frac{\alpha}{\beta})^{1/(n-1)}$. In particular, for $\beta = \pi$, we obtain $K_I(D_\alpha) =
\frac{\pi}{\alpha}, K_O(D_\alpha) = (\frac{\pi}{\alpha})^{1/(n-1)}$, whence $K(D_\alpha) = \frac{\pi}{\alpha}$.
\end{prop}


\begin{rem}
Remarkably, the coefficients of quasiconformality for non-convex domains (i.e. $\pi \leq 2\pi$) are not known.\footnote{at least to the bet of our knowledge}
\end{rem}

Following \cite{car}, we note the following natural generalization of the definition of a wedge:

\begin{defn}
The domain $D_{\alpha k} \subset \mathbb{R}^n, D_{\alpha k} = \{(r,\varphi_1,\dots,\varphi_{n-k
+1},$ $z_{n-k+1},\dots,z_n)\}$, $0 < \varphi_k < \alpha_k\,, 1 \leq k \leq n-\nu-1, \alpha = (\alpha_1,\ldots,\alpha_{n-\nu-1})$, $0 < \alpha_1 \leq 2\pi, 0 < \alpha_2,\ldots,\alpha_{n-\nu-1} \leq \pi$  is called a {\em dihedral wedge of type $\nu$ and angle $\alpha$}.
\end{defn}

\begin{rem}
For $k = n - 2$ we recuperate the classical definition of wedges.
\end{rem}

The numbers that allow us to ascertain whether two domains are quasiconformally equivalent, i.e. that one is the the quasiconformal (therefore homeomorphic) image of the other, and, if so, which is the smallest possible dilatation of such a mapping, are called the {\it coefficients of quasiconformality} (see Appendix 2). We have the following

\begin{prop}[\cite{car}]
The coefficients of quasiconformality for $D_{\alpha k}$ are:

\begin{equation}
K_I(D_{\alpha k}) = \frac{\pi^{n-k-1}}{\alpha_1\cdots \alpha_{n-k-1}}\,\,,
\:\:
K_O(D_{\alpha k}) \geq \left(\frac{\pi^{n-k-1}}{\alpha_1\cdots \alpha_{n-k-1}}\right)^{\frac{1}{n-1}}\,,
\end{equation}

\[K(D_{\alpha k}) = \frac{\pi^{n-k-1}}{\alpha_1\cdots \alpha_{n-k-1}}\,\,.\]
\end{prop}

\begin{cor} \label{cor:conv-poly}
Let $\mathcal{P}$ be a convex polyhedral domain in $\mathbb{R}^n$ and let $m$ denote the number of faces of $\mathcal{P}$. Then we have the following estimates:

\begin{equation}
K_I(\mathcal{P}) \geq \frac{m-n+2}{m-n}\,\,,\:\: K_O(\mathcal{P}) \geq \left(\frac{m-n+2}{m-n}\right)^\frac{1}{n-1},
\end{equation}

\[K(\mathcal{P}) \geq \frac{m-n+2}{m-n}\,\,.\]
\end{cor}

\begin{rem}
Evidently, the same estimates hold for $PL$-smooth convex manifolds.
\end{rem}

Hence, for polyhedra with a very large number of faces, such as encountered in (good) $PL$ approximations of domains $D$ in $\mathbb{R}^3$ (or, more generally, in $\mathbb{R}^n$), having smooth (convex) boundaries, $K(D) \gg 1$. Even without considering approximations and without making appeal to Corollary \ref{cor:conv-poly}, one can easily produce (convex) polyhedra $\mathcal{P}$ that require arbitrarily large
dilatation $K(\mathcal{P})$, by choosing polyhedra with at least one dihedral angle (between
$n$-faces) $\pi/m$, where $m$ is any (arbitrarily large) natural number, and applying Proposition \ref{prop:dist-coef1} directly.\footnote{For a stronger result regarding the nonexistence of isometric embeddings for $PL$ manifolds in dimension $ \geq 3$, see \cite{s}.}

It is interesting to note in this context, that (at least for polyhedral domains in $\mathbb{R}^3$) the ``primary carrier'' of dilatation is the mean curvature $H$ -- see Remark \ref{rem:curvature} above.

\end{enumerate}


\section{
Final comments}

\subsection{Glimmers of hope}
We bring below two different approaches to the embedding problem, that both circumvent the intricacies of the Nash Embedding theorem mentioned in the preceding sections.

\subsubsection{A compromise}
Reviewing the facts above, it is hard not to reach the conclusion that the situation is quite bleak, 
as far as the practical use of Nash's Embedding Theorem is concerned.
However, one may quite 
justifiably 
sustain 
that making appeal to global isometric embeddings in general, and to Nash's Theorem in particular, is to be somewhat overenthusiastic. Indeed, it may be very well claimed,
that one is rarely 
faced, in computer vision, graphics and other related domains, with surfaces (manifolds) globally defined, hence one can restrict himself to {\it local isometric embeddings}. After all, this is the position already adopted (albeit in a different context, where large amounts of data have to be processed) in the widely quoted 
work of Roweis and Saul \cite{RS}.
The method of \cite{TdSL} is also basically local (see, however \cite{DG} for a discussion on its possible globality).

This approach is also augmented by the very first result on isometric embeddings, namely the following theorem of Burstin, Janet and Cartan (see, e.g. \cite{spi5}): 

\begin{thm}[Burstin-Janet-Cartan]
Any (real) analytic manifold $M^n$ can be locally (real) analytically isometrically embedded into $\mathbb{R}^{\frac{n(n+1)}{2}}$.
\end{thm}
%

The problem with the result above is the fact that it requires analyticity. In fact, even if conjectured already by Schlaefly in 1873, the proof of the result above for $\mathcal{C}^\infty$ manifolds is still elusive. 
(It is true, however, that weaker forms of this result -- that is, with higher embedding dimension -- were obtained by Greene \cite{Gre} and Gromov \cite{gr}.)\footnote{The author is not aware of the existence of meaningful, general theorems regarding $\mathcal{C}^k$ manifolds, for $2 \leq k < \infty$.}

\begin{rem} \label{rem:Ck-loc}
As in the global case, for lower differentiability classes, no general results are even possible.
 Indeed, there exist a counterexample, due to Pogorelov \cite{Pog}, of a $\mathcal{C}^{2,1}$ metric on the unit disk $\mathbb{B}^2 = \mathbb{B}^2(0,1) \subset \mathbb{R}^2$, such that there exists no $\mathcal{C}^{2}$ isometric imbedding in $\mathbb{R}^3$ of $\mathbb{B}^2(0,r)$, for any $0 < r < 1$.
(See also \cite{NY1}, \cite{NY2} for some more recent results in this direction.)


\end{rem}

Still, one may argue (rather convincingly) that it is quite common\footnote{even though the author does not subscribe himself to this philosophy}
 in imaging and vision to adopt smooth, even analytic models and consider standard types of approximations (for the manifolds and for various differential operators on these manifolds).


\subsubsection{More dimensions}

Surprisingly, a very effective (at least from the theoretical viewpoint) alternative embedding method follows the quite opposite direction: Instead of reducing the scope of the embedding, one can extend it by adding dimensions. That is, one can embed $M^n$ not in some $\mathbb{R}^N$, but in an infinitely dimensional space, more precisely in $L^\infty(M^n)$ -- the (Banach) space of bounded {\it Borel functions} on $M^n$, endowed with the ``$\sup$'' metric, i.e. $d(f,g) = \sup_{x \in M^n}|f(x) - g(x)|$, for any $f,g \in L^\infty(M^n)$ --
via the {\it Kuratowski Embedding} \cite{Ku}:

\begin{defn}({\it Kuratowski embedding})
Let $M^n$ be a closed Riemannian manifold. Then
\[K:M^n \rightarrow L^\infty(M^n)\,, K(x) = {\rm dist}_x\,,\]
where
\begin{equation}
{\rm dist}_x = {\rm dist}(x,\cdot)\,,
\end{equation}
where ``${\rm dist}$'' denotes the (intrinsic, Riemannian) distance on $M^n$,
is called the {\it Kuratowski embedding} (of $M^n$).
\end{defn}

This method is much more powerful than it would appear at first sight. Indeed, the Kuratowski embedding is an isometry, more precisely we have the following Lemma (see, e.g. \cite{G}):

\begin{lem}
With the notation above, we have
\[d({\rm dist}_x,{\rm dist}_y) = {\rm dist}(x,y)\,.\]
\end{lem}

\begin{rem}
This approach is widely divergent from the Riemennian embedding one adopted in Nash's Theorem. Indeed, the Riemannian and Kuratowski embeddings coincide iff $K(M^n) \subset L^\infty(M^n)$ is a  convex, open subset of an affine linear subspace of dimension $n$.
\end{rem}

On behalf of the Kuratowski embedding, one can remark that, albeit being infinite dimensional, it may be quite advantageous when a functional approach is needed or sought for (e.g. when considering {\it spline functions, wavelets}, etc.).\footnote{A closely related approach is well known to the Imaging and Vision community: Embedding by using the eigenvalues of the Laplacian or of the Green Kernel. (For applications of these methods in the context of Riemannian Geometry, see, e.g. \cite{BBG}, \cite{KaKu}.)} However, usually (and more realistically) the spaces that appear in Computer Science (and even more so in Graphics) are finitely dimensional. Moreover, most people in the said communities find infinitely dimensional spaces
as somewhat of an artifice, highly nonintuitive, and of theoretical value at best.

Fortunately, there exists a finitely dimensional version of the Kuratowski embedding:
Let $X$  be an {\it $\varepsilon$-net}\footnote{Recall that $\varepsilon$-nets are defined as follows:
\begin{defn}
Let $(X,d)$ be a metric space, and let $A \subset X$. $A$ is called an $\varepsilon${\it-net} iff
$d(x,A) \leq \varepsilon,$ for all $x \in X$.
\end{defn}}
in $M^n$, $|X| = m$. Then,
 for small enough $\varepsilon$, $K_X:X \rightarrow l_\infty^m$ is an embedding, where $K_X = K|_X$ -- the restriction of $K$ to X and $l_\infty^m$ denotes the $m$-dimensional Banach space endowed, again, with the  ``$\sup$'' metric: if ${\bf x} = (x_1,
 \ldots,x_p)$, then $||{\bf x}|| = \sup_{i}|x_i|$.

Moreover, we can assure that this ``discrete'' version of the Kuratowski embedding is
{\it bi-lipschitz}, more precisely we have the following result (\cite{G}, \cite{KK}):

\begin{thm}
Let $M^n$ be a compact Riemannian manifold without boundary. Then, for any $C > 0$, there exists a $\varepsilon$-net $X$, where $\varepsilon = \varepsilon(C)$, such that
\begin{equation}
(1-C){\rm dist}(x,y) \leq |K_X(x) - K_X(y)|  \leq {\rm dist}(x,y).
\end{equation}
\end{thm}

Due to the theorem above, the finite dimensional version of the Kuratowski embedding is proves to be very useful in Global Differential Geometry: 
Its use in the study of {\it systoles} was pioneered by Gromov \cite{gr2} (see also \cite{G}, \cite{KK}). It was also employed to prove yet another result of Gromov \cite{gr0} (and Katsuda \cite{Ka}), namely a {\it rigidity} theorem: Informally stated, the theorem in question asserts that if two $n$-dimensional (compact) Riemannian manifolds,  having the same lower bound for their volumes, and upper bounds on diameters and sectional curvatures, are sufficiently close one to each other in the Gromov-Hausdorff (metric) topology\footnote{See, e.g. \cite{Fuk}, \cite{sa0} for a short overview of the notion.}, then they are diffeomorphic.

Since the manifolds usually encountered  in Imaging, Vision, etc., naturally satisfy such bounds, it follows that the result above, as well as the finitely-dimensional Kuratowski embedding in general, are quite relevant for applications in the mentioned fields, in particular for recognition type problems. 


\subsection{A possible solution}

A more realistic approach (both from the theoretical and implementational viewpoints) would be to obtain a Discrete Nash Embedding
Theorem \cite{Li}. A certain amount of confidence in the feasibility of obtaining such a result stems, amongst others, from the existence of discrete versions of the required differential operators and invariants (see, e.g. \cite{br1}, \cite{br2}).

Another geometrization approach stems from the differential geometry of metric spaces (see, e.g. \cite{bm}). By
using a discretization of the (metric) Finsler-Haantjes curvature of curves (\cite{sa}) we can obtain the
embedding, via a proper discretization of the Gauss-Bonnet theorem, of any given metric graph, not into
$\mathbb{R}^n$ (or $l_p$, $\mathbb{H}^n$) but rather into a model space (a model surface, to be more precise) -- see
\cite{sa1}.

Naturally, one expects the two embedding methods considered above to converge and augment each other, particularly
in our purely geometric context, stemming from problems in $PL$ differential geometry, computer
graphics and image processing, where the graphs considered are skeleta of triangulations of manifolds (or of
cell-complexes), and the weights are either edge-weights (i.e. distances between vertices) or/and vertex-weights
(i.e. curvature measures) -- see, e.g. \cite{sa0}, \cite{SAWZ}.

However, both methods, applied in a more general context, rend themselves to various practical implementations, in such areas as multicommodity flows in networks (e.g. for the prediction of informational bottlenecks, discovery of holes, etc.), clustering of statistical data (in particular in bio-informatics -- see, e.g. \cite{sa}) and expanders.


\section*{Acknowledgments}
The author wishes to express his gratitude to 
Konrad Polthier for extending
him a warm invitation and for bringing to his attention the result
of Burago and Zalgaller, and to 
Peter Buser for his wonderful
hospitality and for the numerous stimulating discussions on this
theme and many others that took place during the author's 
visits at
EPFL.

Thanks are also due to Nati Linial for his interest and a number of inspiring discussions and to Mikhail Katz for reminding me of the Kuratowski embedding.


\section*{Appendix 1 -- 
Immersions and Embeddings}

We presume the reader is familiar with the notions of differentiable manifold and tangent space, as well as with basic concepts of topology (for any eventually needed details see \cite{th}), and we recall only the relevant definitions:

\begin{defn}({\it Immersion})
Let $M^m,N^n$ be smooth differentiable manifolds and let $f:M^m \rightarrow N^n$ a differentiable map. If ${\rm rank}f = m$ at each point of $M^m$, then $f$ is called an {\it immersion}.
\end{defn}

\begin{defn}({\it Embedding})
Let $M^m,N^n$ be smooth differentiable manifolds and let $f:M^m \rightarrow N^n$ a differentiable homeomorphism. If $f$ is also an immersion, then it is called an {\it embedding}.
\end{defn}

(Note that, in this case, $m=n$, of course.)

The condition that $f$ be a homeomorphism is very strong and shouldn't considered lightly. Indeed, not even asking that $f$ be injective will suffice, as proven by the (classical) fact that there exists a injective immersion of $\mathbb{R}$ into the ``figure eight'' curve, but this is not an embedding, since it is not a homeomorphism: the image is not even a manifold!\footnote{An even more pathological example can be constructed, where $\mathbb{R}$ is injectively immersed in the (flat) torus $\mathbb{T}^2$ as a dense geodesic (just consider  the image under the covering map of $\mathbb{T}^2$, of a line making an irrational angle with the $Ox$ axis.}

To sum up: The notions of embbeding and immersion are not interchangeable 
-- while any embedding is, in particular, an immersion, the opposite is not true.\footnote{Therefore, even though, for stylistic reasons, one usually tends to avoid repetitions of the same word, one cannot (see e.g. \cite{bbk1}) freely interchange ``immersion'' and ``embedding''...}

If one discards even the differential structure, then topological embeddings (in the sense of that they homeomorphic on their image) are relatively easily obtained 
by

\begin{thm}[Whitney's Theorem]
Every (smooth) manifold of dimension $n$ admits a (smooth) embbeding in $\mathbb{R}^{2n}$ and a (smooth) immersion in $\mathbb{R}^{2n-1}$.
\end{thm}

\begin{rem}
It is easy to prove, for compact manifolds, that an embedding in some finite dimension $N$ exists. It is then progressively (much) harder to discard the compactness restriction and to gradually ``zero in'' to dimension $2n$, via embedding dimensions $(n+1)^2$ and $2n+1$.
\end{rem}

Up to this point we have dealt with classical (``pure'') 
 differential topology, that is the famous ``rubber geometry'' of popularization texts (albeit endowed with some ``smoothness'' - necessary for the ``differential'' part.) At this point, however, we should introduce a bit of ``solid'' geometry, necessary, e.g. for recognition purposes\footnote{Think of the the individualization problems amoebas are faced with...}. The idea is to use a specific ``measuring yard'', for each manifold, that is a {\it Riemannian metric}:

\begin{defn}({\it Riemannian manifolds})
Let $M^n$ be a manifold and let $x = (x_1,\ldots,x_n)$ denote a standard coordinate chart. The Riemannian metric $g$ on $M^n$ is defined by the length element

\[ds^2 = g = \sum_{i=1}^{n}{g_{ij}dx^idx^j}\,,\]
where the functions $g_{ij} = g_{ij}(x_1,\ldots,x_n)$ represent the scalar products of the vector fields $\frac{\partial}{\partial x_i},\frac{\partial}{\partial x_j}$ associated to the given chart:

\[g_{ij} = g\left(\frac{\partial}{\partial x_i},\frac{\partial}{\partial x_j}\right)\,.\]
\end{defn}

Once an infinitesimal distance is introduced, global ones can also be measured (transforming a Riemannian manifold into a ``honest-to-God'' metric space) as follows:

Let $c:[,] \rightarrow M^n$ be a curve. Then its length is given by:

\[{\rm length}(c) = \int_{}^{}||c'(t)||dt\,.\]

(Here is important to recall that $c'(t)$ is just a tangent vector, so $||c'(t)|| = \sqrt{g\left(c'(t),c'(t)\right)}$\,.)

The {\it intrinsic} (or {\it inner}) distance between two points $p,q \in M^n$ is defined as
%
%
\[
d(p,q) = \inf_{c}\{{\rm length}(c)\,|\, {\rm is\; a\; curve
\; of
\; ends\;} p\; {\rm and\;} q\}\,. 
\]

(We have tried here to keep the technical aspects of the definition above to a minimal level; for those insisting on absolute formal correctness, we recommend, for instance, \cite{Ch}.)
%

Of course, the Riemannian metric induces a topology  on $M^n$ (the metric topology), but in fact a much stronger result holds:

\begin{thm}[Palais, \cite{Pa}]
The metric of Riemannian manifolds determines its (smooth) manifold structure.
\end{thm}

Thus, Riemannian manifolds are, a fortiori, smooth manifolds, and the discussion above holds for them as well. However, a much more specific notion of embedding is applicable(relevant) in this case, namely:

\begin{defn}({\it Isometric embedding}) \label{def:IsomEmb}
An embedding $f:M^n \rightarrow \mathbb{R}^N$ is called isometric iff
\begin{equation}
<\frac{\partial f}{\partial x_i},\frac{\partial f}{\partial x_j}> = g_{i,j}\;, 1 \leq i,j \leq n\,,
\end{equation}
where $<\cdot,\cdot>$ is the standard inner product in $\mathbb{R}^N$
\end{defn}

\begin{rem}
Of course one can extend the definition above to include embeddings $f:M^n \rightarrow Q^N$, where $Q$ is a general Riemannian manifold with metric $h$, by imposing, instead of (\ref{def:IsomEmb}), the following condition:
\begin{equation}
<\nabla_if,\nabla_jf>_h = g_{i,j}\;, 1 \leq i,j \leq n\,,
\end{equation}
where $\nabla_if = Df\left(\frac{\partial }{\partial x_i}\right)$
and where $<\cdot,\cdot>_h$ represents the scalar product defined by the Riemannian metric $h$ on $T_{f(x)}(Q)$, ($x = (x_1,\ldots,x_n) \in M^n$).
\end{rem}

%
%
%
%
%
%
%
%

\begin{rem}({\it A common fallacy}) \label{rem:mistake}
Sadly, the notion of isometric embedding (and in particular the result above) are sometimes puzzling even for the professional mathematician (and even, sometimes, for topologists!...) However, things become much simpler if one keeps in mind Definition \ref{def:IsomEmb} and remembers that saying that the intrinsic metric ``equals'' the Euclidean one, means just that the {\it infinitesimal} length element, as defined by the Riemannian metric coincides with that of the {\it infinitesimal} one induced by the ambient Euclidean space. In any case this shouldn't be interpreted as affirming that lengths of curves, as measured on the manifold, equal the Euclidean distance between their ends, as measured in the ambient space $\mathbb{R}^N$ (see Figure below). Therefore, it would be redundant and useless (not to say mistaken) to try and isometrically embed such surfaces in  $\mathbb{R}^N$, for some $N > 3$.\footnote{In fact, the only manifolds for which the embedding and intrinsic metric coincide are precisely the piecewise flat ones -- see \cite{Be}.}

\begin{figure}[htb]
\begin{center}
\includegraphics[scale=0.65]{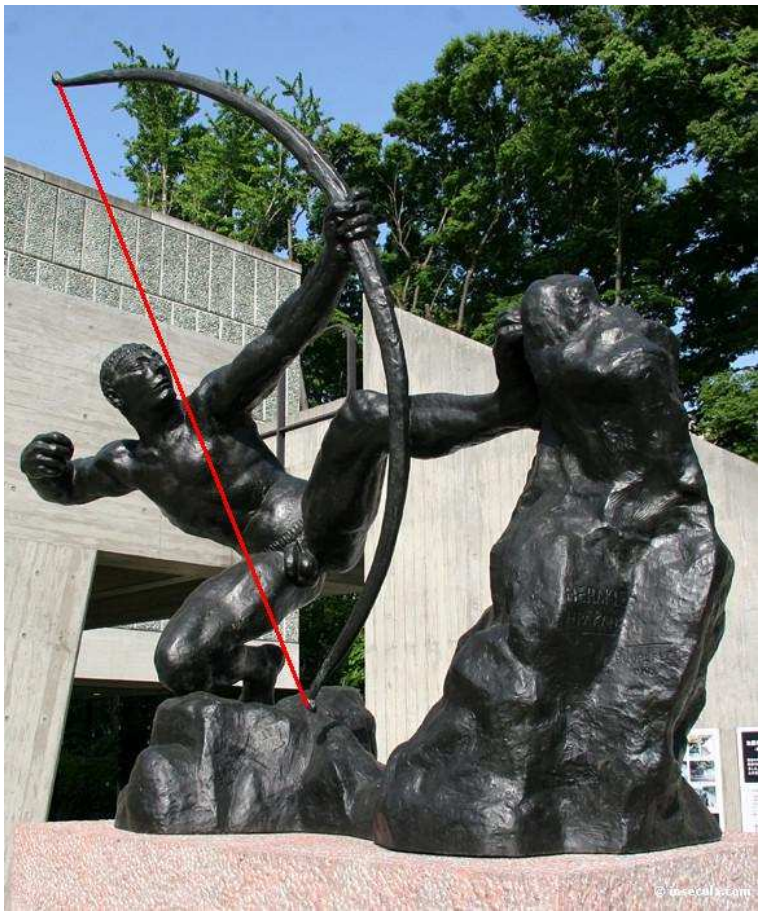}
\end{center}
\caption{Intrinsic vs. ambient distance: The length of the (strait, added) cord never equals the one of the bow, independently of the dimension in which Rodin's bowman resides.}
\end{figure}

On the positive side, surfaces in $\mathbb{R}^3$ are already embedded and inherit, therefore, a Riemannian structure from the ambient space, that is the induced metric defines on the surface a Riemannian metric (even if none was supposed (or given) a priori).

\end{rem}


\section{Appendix 2 -- 
Quasiconformal Mappings}

\begin{defn}({\it Quasiregular and quasiconformal mappings})
Let $D \subseteq \mathbb{R}^n$ be a domain; $n \geq 2$ and let $f: D \rightarrow \mathbb{R}^n$ be a continuous
mapping. $f$ is called
\begin{enumerate}
\item {\it quasiregular} ({\it qr}) iff
\begin{enumerate}
\item
$f$ is locally Lipschitz (and thus differentiable a.e.); 
\\ \hspace*{-0.8cm} and
\item \(0 < |f'(x)|^n \leq KJ_f(x)\), for any \(x \in M^n\);
\end{enumerate}
\;where 
$|f'(x)| = \sup
\raisebox{-0.25cm}{\mbox{\hspace{-0.75cm}\tiny$|h|=1$}}|f'(x)h|$, and where $J_f(x) = detf'(x)$;
\item {\it quasiconformal} ({\it qc}) iff $f:D \rightarrow f(D)$ is a quasiregular homeomorphism;
\end{enumerate}
The smallest number $K$ that satisfies condition (b) above is called the {\it outer dilatation} of
\nolinebreak[4]$f$.
\end{defn}

\begin{rem}
One can extend the definitions above to mappings between oriented, connected
 Riemannian $n$-manifolds, $n \geq 2$, by using coordinate charts (for details see, e.g. \cite{v}).
\end{rem}

\begin{rem}
It follows immediately  from Condition (1)\,(b) above, that {\it qr}-mappings are sense preserving.
\end{rem}


If $f: D \rightarrow \mathbb{R}^n$ is quasiregular, then there exists $K' \geq 1$ such that the following
inequality holds a.e. in $M^n$:

\begin{equation} \label{eq:qr1}
J_f(x) \leq K'\inf_{|h|=1}{|T_xfh|^n}
\end{equation}

By analogy with the outer dilatation we have the following definition:

\begin{defn}({\it $K$-quasiregularity})
The smallest number $K'$ that satisfies inequality (\ref{eq:qr1}) is called the {\it inner dilation} $K_I(f)$ of $f$, and $K(f)
= \max(K_O(f),K_I(f))$ is called the {\it maximal dilatation} of $f$. 
If $K(f) < \infty$ we say that $f$ is called $K$-{\it quasiregular}.
\end{defn}

The dilations are $K(f), K_O(f)$ and $K_I(f)$ are simultaneously finite or infinite. Indeed, the following
inequalities hold: $K_I(f) \leq K_O^{n-1}(f)$ and $K_O(f) \leq K_I^{n-1}(f)$.

\begin{defn}({\it Coefficients of quasiconformality})
Let $D_1,D_2 \subset \mathbb{R}^n$ be domains homeomorphic to each other. The numbers
\begin{equation}
K_O(D_1,D_2) = \inf_f{K_O(f)},
\, K_I(D_1,D_2) = \inf_f{K_I(f)},\, K(D_1,D_2) = \inf_f{K(f)},
\end{equation}
where the infima are taken over all the homeomorphisms $f:D_1 \stackrel{\sim}{\rightarrow} D_2$ are called the {\it outer}, {\it inner} and {\it total coefficient of quasiconformality} of $D_1$ with respect to $D_2$, respectively. If $D_2$ is the unit ball $\mathbb{B}^n$, then the numbers
$K_O(D_1) = K_O(D_1,\mathbb{B}^n)$, $K_I(D_1) = K_I(D_1,\mathbb{B}^n)$, $K(D_1) = K(D_1,\mathbb{B}^n)$ are simply called the (inner, resp. outer, resp. total) coefficients of conformality of $D_1$.
\end{defn}

Again, the numbers $K_O(D_1,D_2), K_I(D_1,D_2)$ and $K(D_1,D_2)$ are simultaneously finite or infinite. However, it is not always guaranteed that there actually exists a homeomorphism $f$ as above, such that $K_I(f) = K_I(D_1,D_2)$ or $K_O(f) = K_O(D_1,D_2)$, nor that if existing, it is unique.  However, in the following important cases such an {\it extremal} mapping  (for $K_I$ or $K_O$) is known to exist:

\begin{thm}[Gehring-V\"{a}is\"{a}l\"{a} \cite{gv}, Gehring \cite{ghe}]
The extremal mappings for $K_I$ and $K_O$ exist if
\begin{enumerate}
\item $D_1$ or $D_2$ is a ball;
\item The boundary of $D_1$, $\partial D_1$ has $k$ components, where $2 \leq k < \infty$;
\item $D_1, D_2$ are tori in $\mathbb{R}^3$.
\end{enumerate}
\end{thm}



\begin{thebibliography}{99}

\bibitem{a}
Andrews, B. {\it Notes on the isometric embedding problem and the
Nash-Moser implicit function theorem}, Proceedings of CMA, Vol. 40,
157-208, 2002.

\bibitem{ban1}
Banchoff, T. A., {\it Critical points and curvature for embedded polyhedra}, J. Differential Geometry {\bf 1},
257-268, 1967.

\bibitem{ban2}
Banchoff, T. A., {\it Critical Points and Curvature for Embedded Polyhedral Surfaces}, Amer. Math. Monthly {\bf 77}, 475-485, 1970.

\bibitem{bea}
Beardon, A. F. {\em A primer of Riemann surfaces}, 
London Mathematical Society Lecture Note Ser. {\bf 78},  Cambridge University Press,
1984.

\bibitem{BBG}
B\'{e}rard, P., Besson, G. and Gallot, S. {\it Embedding Manifolds by their heat kernel}, Geom. Funct. Anal {\bf 4}(4), 373-398, 1994.


\bibitem{Be}
Berger, M. {\it A Panoramic View of Riemannian Geometry}.
Springer-Verlag, Berlin, 2003.

\bibitem{BH}
Bern, M. and Hayes, B. {\it Origami Embedding of Piecewise-Linear Two-Manifolds}, Lecture Notes in Computer Science {\bf 4957}, 617-629, 2008.

\bibitem{bm}
Blumenthal, L. M. and  Menger, K.  {\it Studies in Geometry}, Freeman and Co., 1970.


\bibitem{bbk}
Bronstein, A. M., Bronstein, M. M. and Kimmel, R.: {\it On isometric
embedding of facial surfaces into S3}, Proc. Intl. Conf. on Scale
Space and PDE Methods in Computer Vision, pp. 622-631, 2005.


\bibitem{bbk1}
Bronstein, A. M., Bronstein, M. M. and Kimmel, R.: {\it
Three-dimensional face recognition,   International Journal of
Computer Vision}, 
  64(1):5-30, 2005.
%



\bibitem{br1}
Brooks, R. {\it Reflections on the First Eigenvalue}, Texas Tech Distinguished Lecture Series, vol. 19, 1996.

\bibitem{br2}
Brooks, R. {\it Spectral Geometry and the Cheeger Constant}, J. Friedman(ed), Expanding Graphs, Proc. DIMACS
Workshop, Amer. Math. Soc. 1993.




\bibitem{BPS}
Buchin, K., Pak, I. and Schulz, A. {\it Inflating the cube by shrinking},
In Proc. 23rd European Workshop on Computational Geometry, Pages 46–49, Graz, Austria, March 2007.

\bibitem{BS}
Burago, Yu. D. and Shefel', S. Z. {\it The Geometry of Surfaces in Euclidean Space}, In: Burago, Yu. D. and Zalgaller, V. A. (eds.) ``Geometry III: Theory of Surfaces'', Encyclopedia of Mathematical Sciences {\bf 48}, Springer-Verlag, Berlin, 1992.


\bibitem{bz}
Burago, Yu. D. and Zalgaller, V. A. {\it Isometric piecewise linear
immersions of two-dimensional manifolds with polyhedral metrics into
$\mathbb{R}^3$}, St. Petersburg Math. J., Vol. 7, No. 3, 369-385,
1996.

\bibitem{car}
Caraman, P. {\it n-Dimensional Quasiconformal (QCf) Mappings}, Editura Academiei Rom\^{a}ne, Bucharest, Abacus
Press, Tunbridge Wells Haessner Publishing, Inc., Newfoundland, New Jersey, 1974.


\bibitem{Ch}
Chavel, I. {\it Riemannian geometry -- a modern introduction}, Cambridge Tracts in Mathematics {\bf 108}, Cambridge University Press, 1993.

\bibitem{D'A}
D`Ambra, G. {\it Isometric immersions and induced geometric structures}
In: Jan Slov\'{a}k and Martin \v{C}adek (eds.): Proceedings of the 18th Winter School ``Geometry and
Physics''. Circolo Matematico di Palermo, Palermo, 1999. Rendiconti del Circolo Matematico di
Palermo, Serie II, Supplemento {\bf 59}, 13–23.

\bibitem{doC}
do Carmo, M. P. {\it Differential Geometry of Curves and Surfaces},
Prentice-Hall, Englewood Cliffs, N.J., 1976.

\bibitem{DG}
Donoho, D. L. and Grimes, C. {\it Image Manifolds which are Isometric to Euclidean Space}, Journal of Mathematical Imaging and Vision {\bf 23}, 5-24, 2005.


\bibitem{Fu}
Fu, J. H. G. {\it Convergence of Curvatures in Secant Approximation}, J. Differential Geometry {\bf 37}, 177-190, 1993.

\bibitem{Fuk}
K. Fukaya, {\it Metric Riemannian Geometry}, in Handbook of differential geometry. Vol. II,
189--313, Elsevier/North-Holland, Amsterdam, 2006.

\bibitem{ghe}
Gehring, W. F. {\it Extremal mappings between tori}, Certain problems in mathematics and mechanics, Nauka, Leningrad, 146-152, 1970. (Russian)

\bibitem{gv}
Gehring, W. F. and V\"{a}is\"{a}l\"{a}, J. {\it The coefficients of quasiconformality}, Acta Math. {\bf 114}, pp.
1-70, 1965.

\bibitem{Gre}
Greene, R. {\it Isometric imbeddings of Riemannian and pseudo-Riemannian manifolds}, Mem. Am. Math. Soc. {\bf 97}, 1970.

\bibitem{gr1}
Gromov, M. {\em Isometric immersions and embeddings}, 
Soviet Math. Dokl. 11 (1970), 794-797.

\bibitem{gr0}
Gromov, M. {\it Structures m\'{e}triques pour les vari\'{e}t\'{e}s riemanniennes}, Lafontaine, J. and Pansu, P., eds., Textes Math\'{e}matiques {\bf 1}, CEDIC, Paris, 1981.

\bibitem{gr}
Gromov, M. {\em Partial differential relations}, Springer-Verlag,
Ergeb. der Math. 3 Folge, Bd. 9, Berlin - Heidelberg - New-York,
1986.

\bibitem{gr2}
Gromov, M. {\em Metric Structures for Riemannian and Non-Riemannian
Spaces}, Birkhauser, Second printing, 2001.

\bibitem{gu}
G\"{u}nther, M. {\it On the perturbation problem associated to
isometric embeddings of Riemannian manifolds}, Ann. Global Anal.
Geom. 7 (1989), 69–77.


\bibitem{gu1}
G\"{u}nther, M. {\it Isometric embeddings of Riemannian manifolds},
Proc. ICM Kyoto (1990), 1137–1143.

\bibitem{G}
Guth, L. Notes on Gromov's systolic estimate. Geom. Dedicata 123, 113–129 (2006)


\bibitem{HH}
Han, Q. and Hong, J.-X. {\it Isometric embedding of Riemannian manifolds in Euclidean spaces}, AMS  Math. Surv. {\bf 130}, Providence, R.I., 2006.

\bibitem{HC-V}
Hilbert, D. and Cohn-Vossen, S. {\em Geometry and the Imagination}, Chelsea, 1952.

\bibitem{Ka}
Kalai, G. {\it Rigidity and the lower bound theorem I}, Invent. math. {\bf 88}, 125-151, 1987.

\bibitem{KaKu}
Kasue, A. and Kumura, H. {\it Spectral convergence of Riemannian manifolds}, T\^{o}hoku Math. J. {\bf 46}, 147-179, 1994.

\bibitem{Ka}
Katsuda, A. {\it Gromov's convergence of theorem and its applications}, Nagoya Math. J. {\bf 100}, 11-48, 1985.

\bibitem{ku}
Kuiper, N. {\it On $\mathcal{C}^1$-isometric embeddings. 1} Proc.
Kon. Neder. Akad. Wetensch. A {\bf 58}, 545-556, 1955.


\bibitem{hbssrsr}
Hurdal, M. K., Bowers, P. L., Stephenson, K., Sumners, D. W. L.,
Rehm, K., Schaper. K. and Rottenberg, D. A. {\it Quasi Conformally Flat
Mapping the Human Crebellum}, Medical Image Computing and
Computer-Assisted Intervention -MICCAI'99, (C. Taylor and A.
Colchester. eds), vol. 1679, Springer-Verlag, Berlin, 279-286, 1999.

\bibitem{hs}
Hurdal, M. K. and Stephenson, K. {\it Cortical cartography using the
discrete conformal approach of circle packings}, NeuroImage 23,
119-128, 2004.

\bibitem{KaKu}
Kasue, A. and Kumura, H. {\it Spectral convergence of Riemannian manifolds}, T\^{o}hoku Math. J. {\bf 46}, 147-179, 1994.


\bibitem{KK}
Katz, K. U. and Katz, M. G. {\it Bi-Lipschitz approximation by finite-dimensional imbeddings}, Geom. Dedicata, Online First (DOI 10.1007/s10711-010-9497-4), 2010.


\bibitem{kms}
Kimmel, R. Malladi, R. and Sochen, N.  {\it Images as Embedded Maps
and Minimal Surfaces: Movies, Color, Texture, and Volumetric Medical
Images},  International Journal of Computer Vision,  {\bf 39}(2),
pp. 111-129, 2000.

\bibitem{KBP}
Krat, S., Burago, Yu. D. and Petrunin, Y. D. {\it Approximating Short Maps by PL-Isometries and Arnold's "Can You Make Your Dollar Bigger" Problem},
talk abstract, Forth International Meeting of Origami Science, Mathematics and education, Passadena, 2006.

\bibitem{Ku}
Kuratowski, C. {\it Quelques problemes concernant les espaces
metriques non-separables}, Fund. Math. {\bf 25},
534-545, 1935.

\bibitem{LSE}
Lev, R., Saucan, E. and Elber, G. {\it Curvature Estimation over Smooth Polygonal Meshes using The Half Tube Formula},
Lecture Notes in Computer Science, 
{\bf 4647}, 275-289, 
2007.

\bibitem{Li}
Linial, N. {\it personal communication}.

\bibitem{NY1}
Nadirashvili, N. and Yuan, Y. {\it Counterexamples for Local Isometric Embedding}, arXiv:math/0208127v1 [math.DG], 2002.

\bibitem{NY2}
Nadirashvili, N. and Yuan, Y. {\it Improving Pogorelov's isometric embedding counterexample}, Calculus of Variations and Partial Differential Equations {\bf 32}(3), 319-323, 2008.



\bibitem{na1}
Nash, J. {\it $\mathcal{C}^1$ isometric imbeddings}, Ann. of Math.
{\bf 60}, 383–396, 1954.

\bibitem{na}
Nash, J. {\it The embedding problem for Riemannian manifolds}, Ann.
of Math. (2) {\bf 63}, 20-63, 1956.

\bibitem{na3}
Nash, J., {\it Analyticity of the solutions of implicit function problem with analytic data}, Ann. of Math. {\bf 84}, 345-355, 1966.

\bibitem{Pak1}
Pak, I. {\it Inflating polyhedral surfaces}, preprint; available at http://math.mit.edu/~pak

\bibitem{pak2}
Pak, I. {\it Inflating the cube without stretching}, Amer. Math. Monthly 115 (2008), 443–445;
arXiv:math.MG/0607754.

\bibitem{Pa}
Palais, R. S. {\it On the differentiability of isometries}, Proc. Amer. Math. Soc. {\bf 8}, 805-807, 1957.

\bibitem{Pog}
Pogorelov, A. V. {\it An example of a two-dimensional Riemannian metric that
does not admit a local realization in $E_3$}, Dokl. Akad. Nauk SSSR {\bf 198},
42-43, 1971.

\bibitem{po}
Polthier, K. {\it personal communication}.

\bibitem{Pol}
Polya, G.
{\it An elementary analogue of the Gauss-Bonnet theorem}, Amer. Math. Monthly {\bf 61}, 601-603, 1954.

\bibitem{RS}
Roweis S. T. and Saul, L. K. {\it Nonlinear Dimensionality Reduction by Locally Linear Embedding},  Science
{\bf 290}(5500), 2323 - 2326, 2000.

\bibitem{s}
Saucan, E. {\it Triangulation, Differential Geometry and Quasimeromorphic Mappings}, preprint.

\bibitem{sa}
Saucan, E. and  Appleboim, E. {\it Curvature Based Clustering for DNA Microarray Data Analysis}, with Eli
Appleboim, Lecture Notes in Computer Science, IbPRIA 2005, {\bf  3523}, pp. 405-412, Springer-Verlag, 2005.

\bibitem{sa0}
Saucan, E. and Appleboim, E. {\it Metric Methods in Surface Triangulation}, 
 Lecture Notes in Computer Science, {\bf 5654}, 335-355, 2009.

\bibitem{sa1}
Saucan, E. and  Appleboim, E. {\it Can  One  See the Shape of a Network? -- Geometric Viewpoint of Information
Flow}, in preparation.


\bibitem{SAWZ}
Saucan, E. Appleboim, E., Wolansky, G.  and  Zeevi, Y. Y. {\it Combinatorial Ricci Curvature and Laplacians for Image
Processing},  Proceedings of CISP'09, Vol. 2, 992-997, 2009.

\bibitem{spi5}
Spivak, M., {\it A Comprehensive Introduction to Differential Geometry, volume V}, Publish or Perish, Boston, MA,
1975.

\bibitem{St}
Stone, D. A.,
{\it Sectional curvature in piecewise linear manifolds}, Bull. Am. Math. Soc. {\bf 79}(5), 1060-1063, 1973.

\bibitem{Ta}
Tasmuratov, S. S. {\it The bending of a polygon into a polyhedron with a given boundary},
Sib. Math. J. 15, 947-953, 1974.

\bibitem{TdSL}
Tenenbaum, J. B., de Silva, V. and Langford, J. C. {\it A global geometric
framework for nonlinear dimensionality reduction}, Science,
Vol. 290, No. 5500, pp. 2319-2323, 2000.

\bibitem{th}
Thurston, W. Three-Dimensional Geometry and Topology
(Levy S, Ed.). Princeton: Princeton University Press 1997.

\bibitem{to}
Tompkins, C. {\it Isometric embedding of flat manifolds in Euclidean space},  Duke Math. J.  5, no. 1 (1939),
58-61.

\bibitem{v}
V\"{a}isal\"{a}, J. {\it Lectures on $n$-dimensional quasiconformal mappings}, Lecture Notes in Mathematics 229,
Springer-Verlag, Berlin - Heidelberg - New-York, 1971.

\bibitem{Za}
Zalgaller, V. {\it Isometric imbedding of polyhedra}, Dokl. Akad. Nauk. S.S.S.R. {\bf 123}, 599-601, 1958.

\bibitem{Za}
Zamfirescu, T. {\it Acute triangulations: a short survey}, Proc. Sixth National Conference
of S.S.M.R., Sibiu, Romania, 2002, 9-17.




\end{thebibliography}
\end{document}